\documentclass[conference]{IEEEtran}
\IEEEoverridecommandlockouts

\newif\ifdoubleblind
\doubleblindfalse

\newif\ifralshort
\ralshortfalse

\usepackage{cite}
\usepackage{amsmath,amssymb,amsfonts}
\usepackage{graphicx}
\PassOptionsToPackage{hyphens}{url}
\usepackage{hyperref}
\usepackage{textcomp}
\usepackage{xcolor}
\usepackage{booktabs}
\usepackage{array}
\usepackage[letterpaper,left=0.75in,right=0.75in,top=0.75in,bottom=0.75in]{geometry}
\def\IEEEtitletopspaceextra{0.25in}

\graphicspath{{figures/}{hardware_design/automotive-cable-harness/media/}{hardware_design/gearbox-assembly/media/assembly/images/}}

\def\BibTeX{{\rm B\kern-.05em{\sc i\kern-.025em b}\kern-.08em
    T\kern-.1667em\lower.7ex\hbox{E}\kern-.125emX}}

\begin{document}

\title{Industrial Dexterity Benchmark: A Hardware–Software Benchmarking Platform for Industrial Dexterous Manipulation}

% --- Author list: anonymized under double-blind review. ---
\ifdoubleblind
% \author{Anonymous Author(s)}
\else
\author{Honglu He$^{\dagger}$, Jacob Laufer$^{\dagger}$, Zhiwu Zheng, David Elkan-gonzalez, Raman Goyal, Xinyi Li, Su Lu, \\ Mishek Musa, Berke Saat, Nicolas Tan, and Colm Prendergast
\thanks{All authors are with Analog Devices, Inc., USA (e-mail:
\{honglu.he, jacob.laufer, zhiwu.zheng, david.elkan-gonzalez, raman.goyal,
matteo.grimaldi, xinyi.li, su.lu, mishek.musa, berke.saat, nicolas.tan,
colm.prendergast\}@analog.com).}
\thanks{$^{\dagger}$These authors contributed equally to this work.}
}
\fi

\maketitle
\pagestyle{empty}
\thispagestyle{empty}

\begin{abstract}
Dexterous manipulation remains a critical bottleneck in industrial
automation; tasks such as cable routing, connector insertion, and
precision assembly still rely heavily on manual labor despite decades
of robotics research. This work presents a progression from
classical, modular robotics pipelines toward an end-to-end multimodal
imitation-learning framework for industrial dexterous manipulation.
As a part of this work, we introduce three key contributions: a set of
\textit{Industrial Dexterity Benchmark} (IDB) boards aimed to mimic 
datacenter cable management, automotive cable harnesses, and gearbox assembly tasks; a scalable imitation learning
framework; and a multimodal diffusion-based
policy framework (\textit{AG-iDP3}) that creates models fusing RGB images, point clouds, joint positions, and wrist-frame wrench data. Focusing on the datacenter cable
manipulation board, we evaluate the performance of a task involving cleaning a single cable over variations of an end-to-end AI policy using 48 trials per configuration. 
The best performing configuration, a multimodal
expansion Diffusion Policy (DP), includes a multi-view RGB image source passed through an R3M encoder and reaches a 76\% grasp and insert combined task success rate. This performance marks a significant improvement over the 36\% observed from the
single-camera RGB DP baseline. Each of the tested configurations requires only $\sim$100
teleoperated demonstrations per task phase. 
\ifralshort
These results show that fusing multimodal sensing including multi-view RGB, point clouds, and wrist-frame wrench with a compliant behavior-tree deployment substantially improves task success over a single-camera baseline while requiring only modest, on-site data collection, pointing toward scalable imitation learning for high up-time industrial environments.
\else
These results indicate that the correct
learned policy can outperform classical vision and control robotic methods in robustness,
generalization, and deployment efficiency, justifying a shift toward
scalable robotic automation for high up-time industrial environments.
\fi
\end{abstract}


\section{Introduction}
Dexterous manipulation such as cable routing,
connector insertion, and precision assembly in industrial settings remains largely manual
\cite{trommnau2019harness,navas2022harness},
requiring tight coordination between perception, planning, and
compliant control. Among the most demanding instances is datacenter
cable management, where up-time requirements exceed 99.99\%
\cite{barroso2018datacenter} and dense rack layouts and thermal
constraints restrict physical access. Currently, in these industrial datacenters, human intervention is common, especially for cable installation, routing, and replacement
\cite{mogul2023deployability,zhao2019rewiring}.
This manual requirement exposes a clear gap between the operational demands
of modern infrastructure and the capabilities of traditional robotic
automation in certain industrial settings.

\begin{figure}[htbp]
\centerline{\includegraphics[width=0.95\columnwidth]{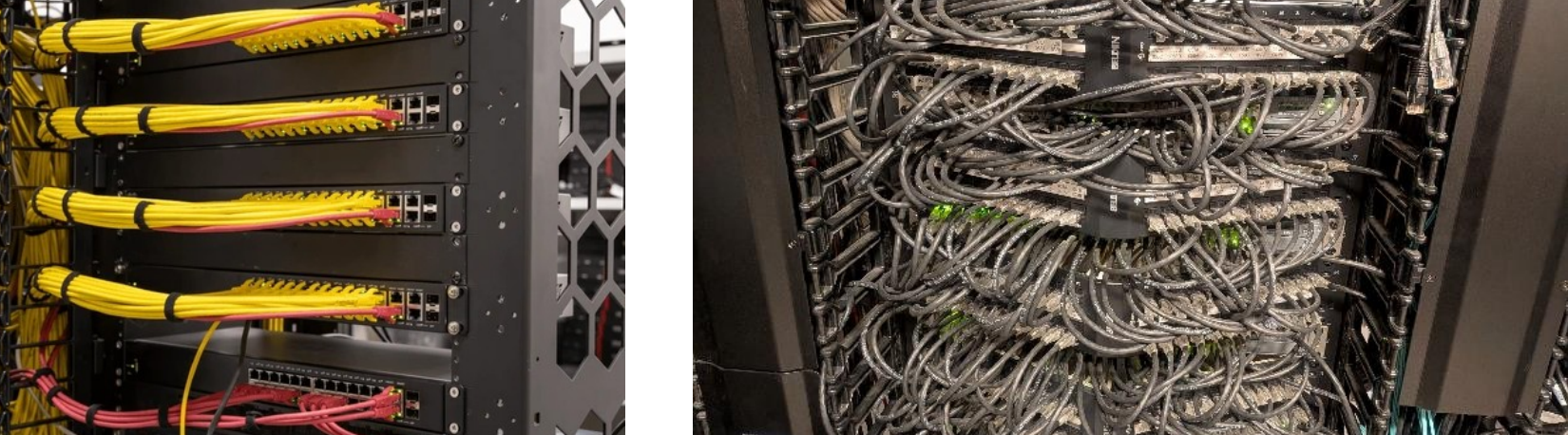}}
\caption{The operational reality this work targets: densely populated
rack switches in production datacenters where up-time targets exceed
99.99\%, ports are tightly spaced, and routine cable swaps
demand contact-aware manipulation that does not disturb neighboring
connectors.}
\label{fig:motivation}
\end{figure}

Fig.~\ref{fig:motivation} shows a real-world example of a datacenter challenge that this work focuses on. The images show densely populated rack switches with sub-millimeter cable separation and routing at sharp angles. In environments like
these, even routine maintenance, like a single connector cleaning or cable swap, requires
careful coordination of perception, planning, and contact-aware
control to complete the task without undesirable disturbances to the environment.

This work originated from efforts to solve National Institute of
Standards and Technology (NIST) Assembly Task Board~\#4 (ATB4)
\cite{nist_taskboard} using classical computer vision paired with classical control techniques.
That solution worked under controlled conditions but proved brittle and
difficult to scale. 
These findings prompted us to develop our own benchmarking standard to better mimic the variability and challenges of the real-world datacenter and switch to using end-to-end learned policies to solve the benchmarks in a more flexible and scalable manner.

\subsection{Contributions and Outline}
We outline three key contributions from our work:
\begin{enumerate}
    \item The \textit{Industrial Dexterity Benchmark (IDB) Boards} (Section~\ref{sec:idb}), inspired by the NIST Assembly Task Board (ATB) set, comprised of three custom benchmarking boards 
    tailored to datacenter fiber switches (IDB Board~\#1), automotive cable
    harnesses (IDB Board~\#2), and planetary gearbox assembly (IDB
    Board~\#3);
    \item An imitation-learning framework (Section~\ref{sec:dagros}), a ROS2-based framework that ties together teleoperation, sensor and sub-component integration, data collection, and real-time robotic arm control;
    \item \textit{AG-iDP3} (Section~\ref{sec:agidp3}), a framework for multimodal diffusion-based policies that are deployed and chained together using a PyTrees-based behavior tree architecture that takes RGB images, point clouds, joint state,
    and robot wrist wrench as inputs and outputs robot actions.
\end{enumerate}

Section~\ref{sec:bt} outlines the evaluator nodes which are used to determine the completeness of each learned task in the behavior tree orchestration.
\ifralshort
Section~\ref{sec:related} reviews related work in standardized assembly benchmarks, diffusion policies, and multimodal perception for manipulation.
\else
Section~\ref{sec:related} reviews related work and the classical proof of concept that motivated our transition away from classical computer vision to learned end-to-end policies.
\fi
Section~\ref{sec:results} reports results comparing six different AI policies developed based on hardware and sensor configurations, all aimed at solving the cable-cleaning task on the IDB Board~\#1. 
The best performing configuration, a multimodal expansion Diffusion Policy (DP), achieves a 76\% combined grasp\,$+$\,insert success
rate compared with 36\% for the single-camera RGB DP baseline. Each trained policy takes no more than $\sim$100 teleoperated demonstrations per task phase.
\section{Related Work}\label{sec:related}

\subsection{Standardized Assembly Benchmarks}
The NIST ATB boards provide a low-cost, reproducible
environment for evaluating robotic systems on industry-relevant tasks. 
The original ATB1-ATB3 boards target small-part assembly
\cite{kimble2020nist_taskboards} while the ATB4 board targets a 
deformable-objects extension with cable harness manipulation \cite{nist_taskboard}.
A recent survey provides a comprehensive review of the state of deformable-object
manipulation \cite{zhu2022deformable} and
industrial cable-routing demonstrations
\cite{hong2024datacenter} and confirm that these tasks are feasible for
modern robotics yet current robotic systems remain far from production ready.
% When designing our IDB board set, we follow the same motivation and goals as NIST but focus our 
% boards on mimicking industrial conditions for datacenter fiber switches, automotive cable harnesses and planetary gearbox assembly tasks.

\ifralshort

\else
\subsection{Classical Proof of Concept on NIST ATB4}\label{sec:classical}
In our initial proof-of-concept work, we use an xArm6 robot equipped with a wrist camera and a force/torque sensor to
perform cable harness insertion on the ATB4 board. This task, proposed by NIST,
quantifies a robot's ability to insert connectors based on a wiring diagram.

\begin{figure}[htbp]
\centerline{\includegraphics[width=0.95\columnwidth]{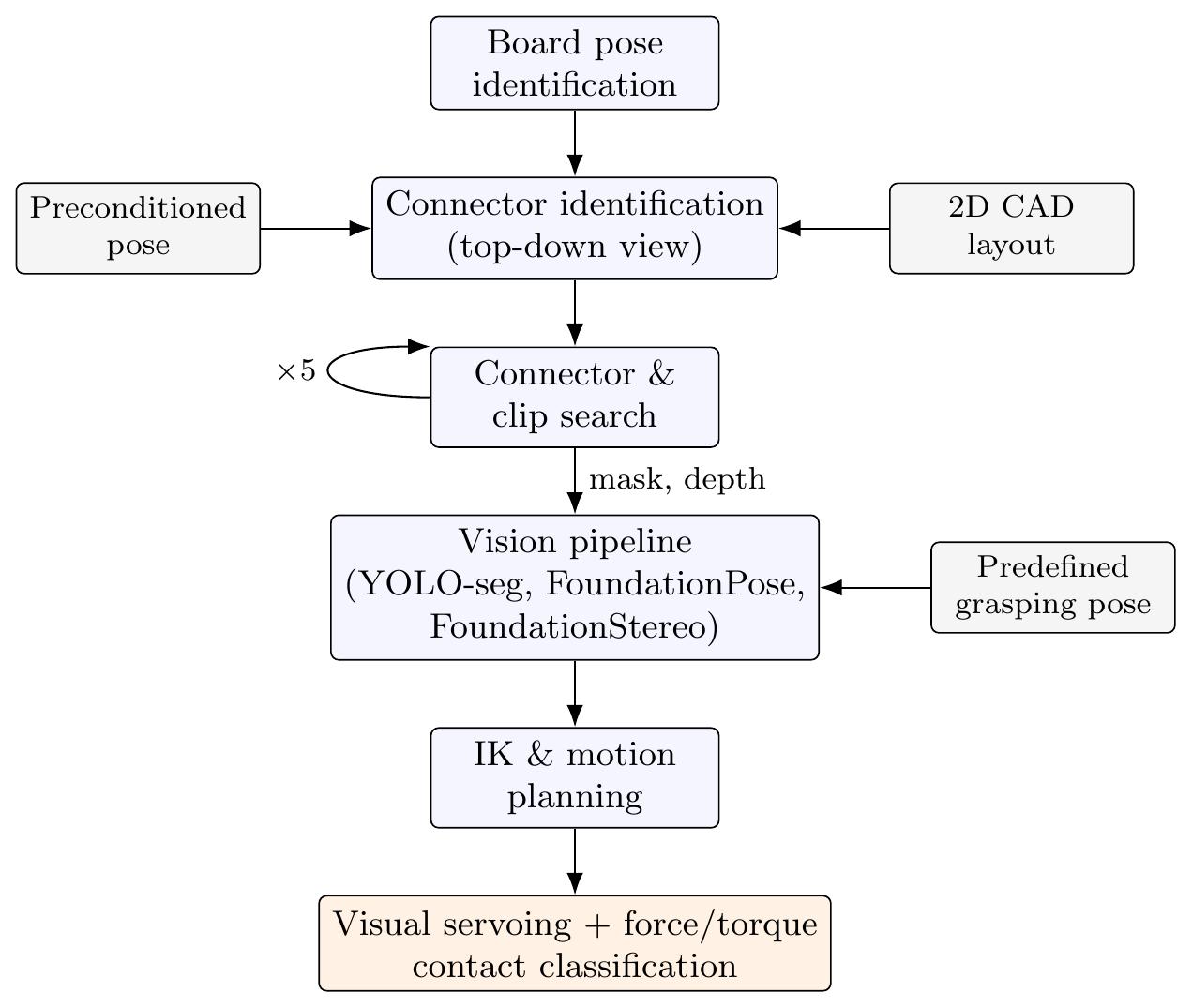}}
\caption{Classical vision pipeline used to solve the
six-connector harness insertion challenge on NIST ATB4.}
\label{fig:vision}
\end{figure}

Fig.~\ref{fig:vision} depicts the six-step automation pipeline we used in our classical robotic solution to the NIST ATB4 board cable insertion challenge. A \emph{board pose identification} module first localizes the
board in the workspace using the fiducial markers; then, given the anchor, a top-down
\emph{connector identification} depth image is taken and matched to the connector 2-D CAD layout in the preconditioned pose\footnote{A connector is in a \emph{preconditioned pose} when it hangs within 5\,cm of its receptacle and the cable lies within 45$^\circ$ of the gravity vector in the connector's vertical plane, ensuring a feasible robot grasp. Infeasible poses are not considered in this work.} and, based on this matching, candidate receptacles are identified. A circular
\emph{connector and clip search} stage repeats up to five times, each time 
refining the connector pose estimate which, along with the connector and clip masks, is fed into the
\emph{vision pipeline} with the depth images.  

The vision pipeline is composed of You Only Look Once (YOLO) for
segmentation \cite{yolo_seg}, the FoundationStereo \cite{wen2025foundationstereo} foundational model for
depth perception, and FoundationPose \cite{wen2024foundationpose} for 6-DOF object pose recognition. The output of the
vision pipeline informs the inverse kinematic solution and motion planning stage to align the connector
directly above and in axis with the receptacle. Next a visual-servoing step further 
aligns the clip of the connector to its mating latch on the receptacle to increase the likelihood of a successful first insertion. 
Finally, as the connector is lowered into the receptacle, a force/torque classification indicates a
successful insertion and on failed insertion, a closed-loop force/torque spiral search algorithm 
ensures a successful connection when the vision-only pipeline fails.

Fig.~\ref{fig:poc_vision_flow} illustrates the key intermediate
vision pipeline steps used to obtain the 6-DOF connector pose. FoundationPose registers
a 3-D CAD model of the connector (Fig.~\ref{fig:poc_vision_flow}a) against the scene. 
To provide the necessary metric 3-D context for accurate pose estimation, FoundationStereo 
reconstructs a dense depth image from the RealSense camera's stereo image pair
(Fig.~\ref{fig:poc_vision_flow}b). The resulting 6-DOF pose
estimates are projected as bounding boxes overlaid on the
detected connectors (Fig.~\ref{fig:poc_vision_flow}c-d), which are
directly fed into the motion planner for robotic grasping.

\begin{figure}[htbp]
\centering
\begin{tabular}{@{}cc@{}}
\includegraphics[width=0.18\columnwidth]{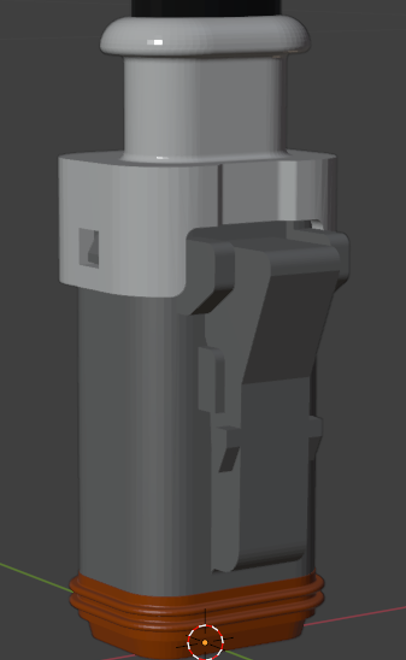} &
\includegraphics[width=0.4\columnwidth]{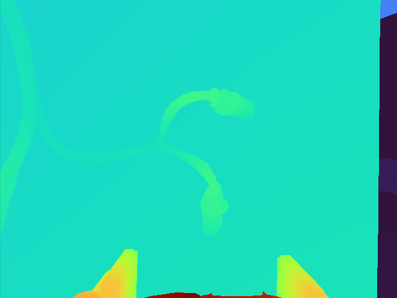} \\
{\footnotesize (a) CAD model input to FoundationPose} &
{\footnotesize (b) FoundationStereo Depth} \\[6pt]
\includegraphics[width=0.4\columnwidth]{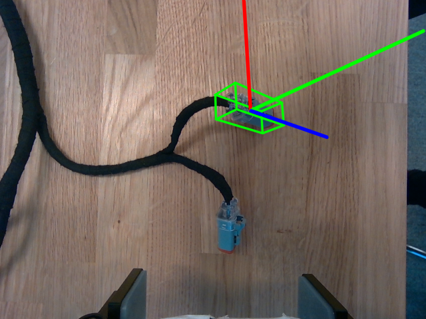} &
\includegraphics[width=0.4\columnwidth]{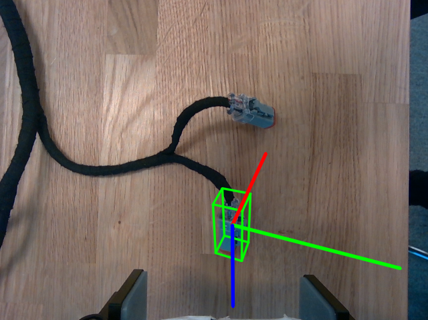} \\
{\footnotesize (c) Connector 1 Pose Estimate} &
{\footnotesize (d) Connector 2 Pose Estimate} \\
\end{tabular}
\caption{Intermediate representations in the classical vision pipeline.
(a)~Textured CAD model provided as the reference mesh to FoundationPose.
(b)~Dense depth map reconstructed from RealSense stereo IR pairs via
FoundationStereo. (c,\,d)~Projected 6-DOF bounding boxes produced by
FoundationPose, overlaid on detected connectors in the scene.}
\label{fig:poc_vision_flow}
\end{figure}

While functional, the classical vision and motion pipeline exhibits several practical
challenges: sensitivity to lighting conditions and environment; difficulty 
in estimating in-hand connector pose; and significant time needed for recalibrating, 
reprogramming, and redesigning the system for even minor task space changes 
(e.g., a similar connector with a different number of contacts). 
These limitations exposed a scalability and robustness gap when classical pipelines are applied
to real-world industrial environments, motivating the move toward
end-to-end learned policies.

\fi

\subsection{Diffusion Policies for Manipulation}
Diffusion Policy \cite{chi2023diffusion} introduced
denoising-diffusion action models \cite{ho2020ddpm} for visuomotor
control; DP3 \cite{ze2024dp3} extended this to a 3-D point-cloud
observation, and iDP3 \cite{ze2025idp3} improved sample efficiency and
stability. In the dexterous manipulation domain, DexDiffuser
\cite{weng2024dexdiffuser} applied diffusion models to generate
dexterous grasps for multi-fingered hands,
\cite{chi2024universal} demonstrated universal manipulation via
diffusion-based interfaces that generalize across end-effector
morphologies, and DexHandDiff \cite{liang2025dexhanddiff} proposed
interaction-aware diffusion planning for adaptive dexterous
manipulation. DexCap \cite{wang2024dexcap} showed that human hand
motion mapping to robotic hand actuation can supply demonstrations for dexterous 
diffusion policies, while Wu et al.\ \cite{wu2024dexpregrasp} combined diffusion
policies with pre-grasp manipulation to reorient objects before grasping.

Despite the many recent advances in diffusion policies for robotic manipulation, 
the existing research shares several limitations. 
First, previous work typically uses either robot joint state data alone
\cite{wu2024dexpregrasp} or combines the joint states with a single image modality such as RGB 
images \cite{chi2023diffusion,chi2024universal} or point clouds
\cite{ze2024dp3,ze2025idp3,weng2024dexdiffuser}, but does not evaluate the effects of
combining different image modalities in a single diffusion policy. 
Second, the previous work is typically conducted either in
simulation \cite{weng2024dexdiffuser,liang2025dexhanddiff} or using
tabletop pick-and-place tasks with large tolerances, leaving
open the question of whether the methods transfer to tight-clearance
industrial assembly. Third, none of the above works incorporate
force/torque sensing as input to the diffusion policy, which is critical
for contact-rich insertion where visual occlusion is unavoidable.

AG-iDP3 addresses these gaps by fusing robot joint states, 
RGB images, 3-D point clouds, and wrist wrench into a unified diffusion policy. 
By incorporating a modality-gating mechanism, the architecture can selectively mask 
sensor inputs during the training phase, enabling us to conduct ablation studies across six distinct sensor configurations. In addition, the policies proposed are trained on real
demonstrations from solving the cable-cleaning task on IDB Board~\#1.

\subsection{Visual Representations and Perception for Manipulation}
R3M \cite{nair2022r3m} provides a generic visual representation for
robot manipulation; we use it as the RGB encoder in AG-iDP3 and
fine-tune it on our collected task demonstrations. For the point-cloud encoding,
we use the same lightweight PointNet \cite{qi2017pointnet} architecture on downsampled data as the original DP3 \cite{ze2024dp3} and iDP3 \cite{ze2025idp3} papers. Beyond vision, proprioceptive signals, including joint
angles and wrist wrench, are passed through unencoded and concatenated directly with the visual features as input to the diffusion U-Net, providing the policy with explicit state and
contact information to complement the learned visual features.

\section{Industrial Dexterity Benchmarking (IDB)}\label{sec:idb}
To enable systematic evaluation and development of our end-to-end policies on industrially relevant tasks, we designed a family of
benchmarking platforms that target distinct industrial manipulation
challenges. The IDB program comprises three benchmark board designs: 
(1)~Datacenter Cable Manipulation,
(2)~Automotive Cable Harness, and (3)~Gearbox Assembly. All three
follow the NIST assembly task board philosophy~\cite{nist_taskboard}; easily fabricated using 
3-D printing, laser cutting, and off-the-shelf hardware with a total cost in the low hundreds of USD. The CAD drawings, bill of materials (BOM), fabrication instructions, and scoring metrics are published online.
\ifdoubleblind
\footnote{removed link for anonymous review.}.
\else
\fi
In this paper we report results exclusively on the
Datacenter Cable Manipulation board; the remaining two designs
are described here for completeness and will be the subject of future
work.

\subsection{Design 1: Datacenter Cable Manipulation (IDB Board \#1)}
\begin{figure}[htbp]
\centerline{\includegraphics[width=0.95\columnwidth]{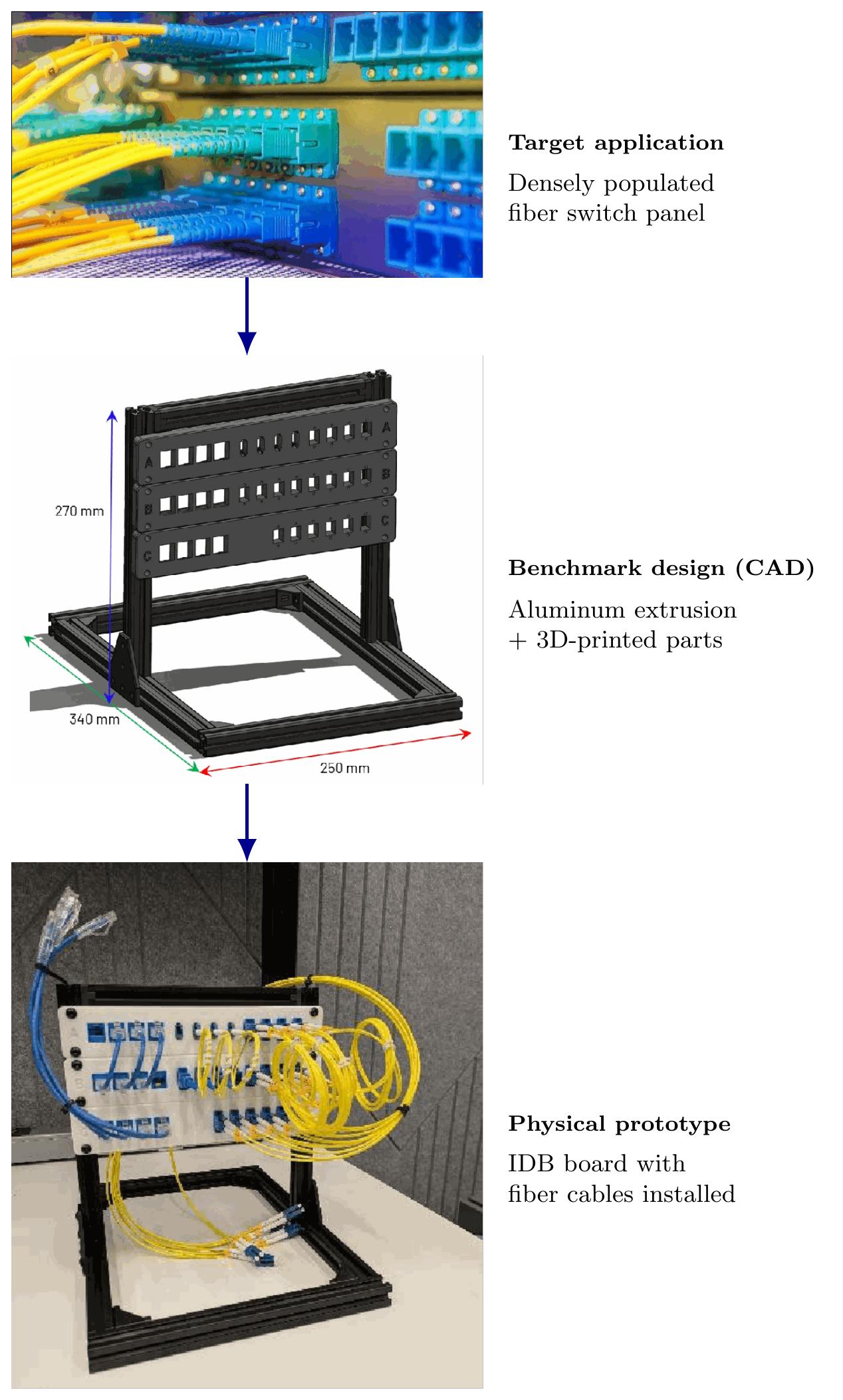}}
\caption{IDB Board~\#1: datacenter cable-management benchmark. Top to
bottom: the target application (densely populated fiber rack), the
initial benchmark design (aluminum extrusion + 3-D-printed parts,
dimensioned), and the final assembly of the board used in this study.}
\label{fig:idb}
\end{figure}

The datacenter benchmark board (IDB Board~\#1) is designed to replicate the dense connector
spacing, constrained cable routing paths, and limited workspace
accessibility characteristic of populated rack switches in industrial datacenters. The board is
assembled from aluminum extrusion ($\sim$270~mm tall, $250\times340$~mm
footprint) with three 3-D-printed patch panels (Panels~A, B, and C)
housing ports for RJ45, SC simplex, and LC duplex fiber connections. The board also contains 
a 3-D-printed circular pad to mimic a connector contact cleaner used in industry for routine maintenance and troubleshooting. 
Fig.~\ref{fig:idb} traces the motivation and design of the IDB Board~\#1.

Two manipulation tasks are defined: \emph{cable port
swap}\footnote{Unplug a cable from one port and re-insert it into an adjacent
port.} and \emph{fiber cable maintenance}\footnote{Unplug a cable from one port, brush the end of the connector on the cleaning pad, and re-insert the cable back to the original port.}. Scoring is a binary
pass/fail for each phase and for a successful task attempt the cable must be fully seated in the
target port without damaging or compromising the connection of any other cable on the board. The level of difficulty
of the tasks can be modulated by changing the number of cables which populate the 
board at the time of task execution.

\subsection{Design 2: Automotive Cable Harness (IDB Board \#2)}
\begin{figure}[htbp]
\centerline{\includegraphics[width=0.66\columnwidth]{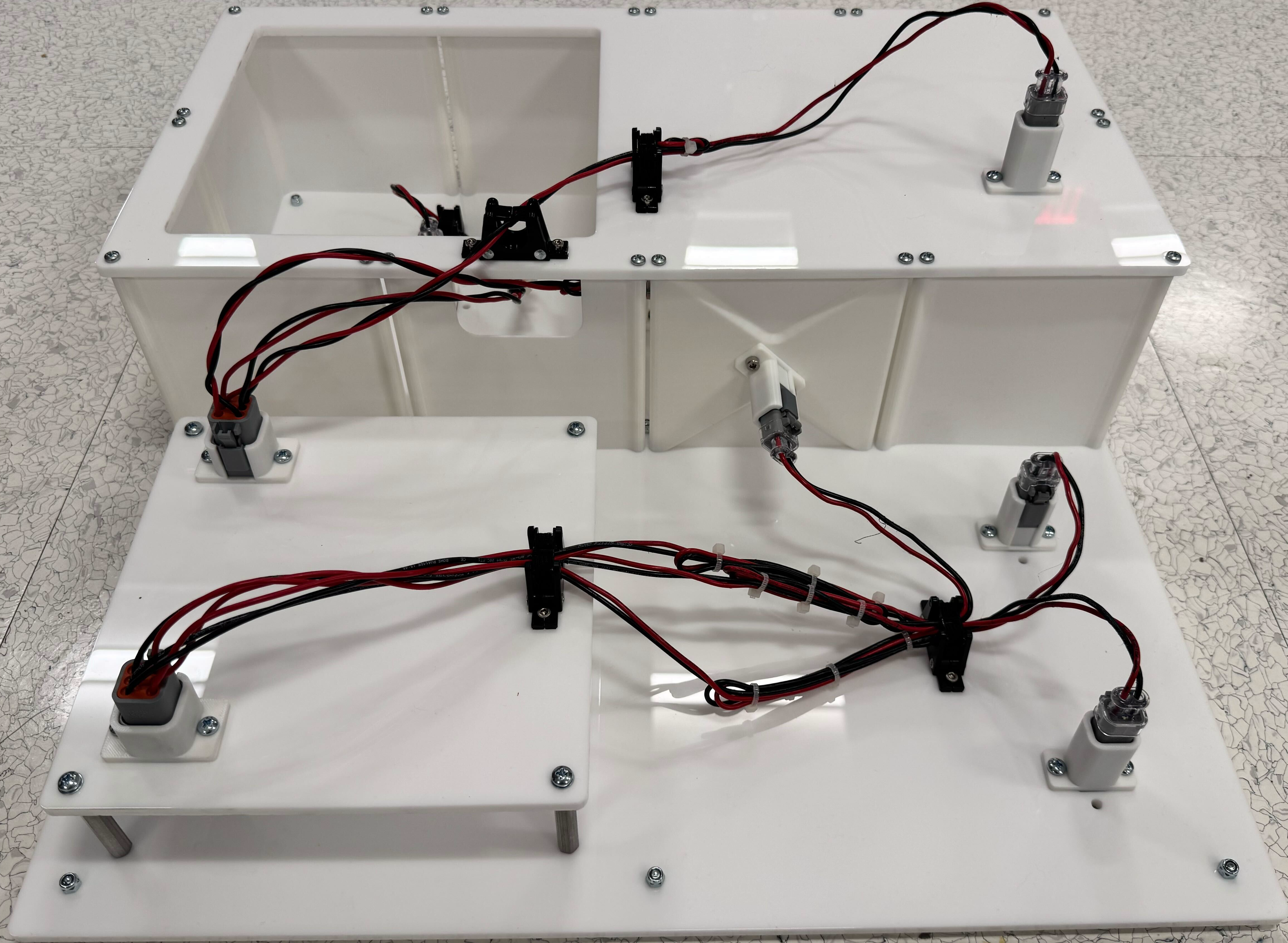}}
\caption{Automotive cable harness benchmarking board (IDB Board~\#2).}
\label{fig:idb_harness}
\end{figure}

The automotive cable harness benchmark (IDB Board~\#2) tests a robot's
ability to manipulate flexible cables in constrained, partially occluded
spaces and to insert automotive-grade connectors into the correct receptacle with precise alignment. IDB Board \#2 
consists of three laser-cut acrylic panels (base, mid, and top)
connected by standoffs, surrounded by 3-D-printed walls that create an
enclosed hard-to-reach region that mimics the constrained electrical connections of automotive final assembly. 
The task is to route two custom 6-pin cable harnesses\footnote{One end of the cable harness has a 6-pin Deutsch AT connector, branching into
three cables terminated with 2-pin Deutsch AT connectors on the other end.} through bungee routing
clips and connect them to correct receptacles.
Fig.~\ref{fig:idb_harness} shows the assembled board.

The benchmark includes eight connectors (two 6-pin, six 2-pin) distributed at varying heights, orientations, and levels of
occlusion. The 2-pin connectors are equipped with LEDs that illuminate
upon successful electrical connection, providing unambiguous pass/fail
feedback. Receptacles in the enclosed area have multiple mounting positions 
allowing users to vary the level of difficulty of the task using the same board. 
An advanced routing option requires threading a cable through a narrow cutout in the wall, 
further stressing spatial awareness and, likely, robotic bimanual coordination.
Scoring evaluates both connection success (eight connectors, LED-verified)
and cable routing compliance (whether cables pass through the prescribed
routing clips). 
% No time limit is imposed and no penalties are assessed
% for re-attempts, enabling clean separation of capability from speed.

\subsection{Design 3: Gearbox Assembly (IDB Board \#3)}
\begin{figure}[htbp]
\centerline{\includegraphics[width=0.66\columnwidth]{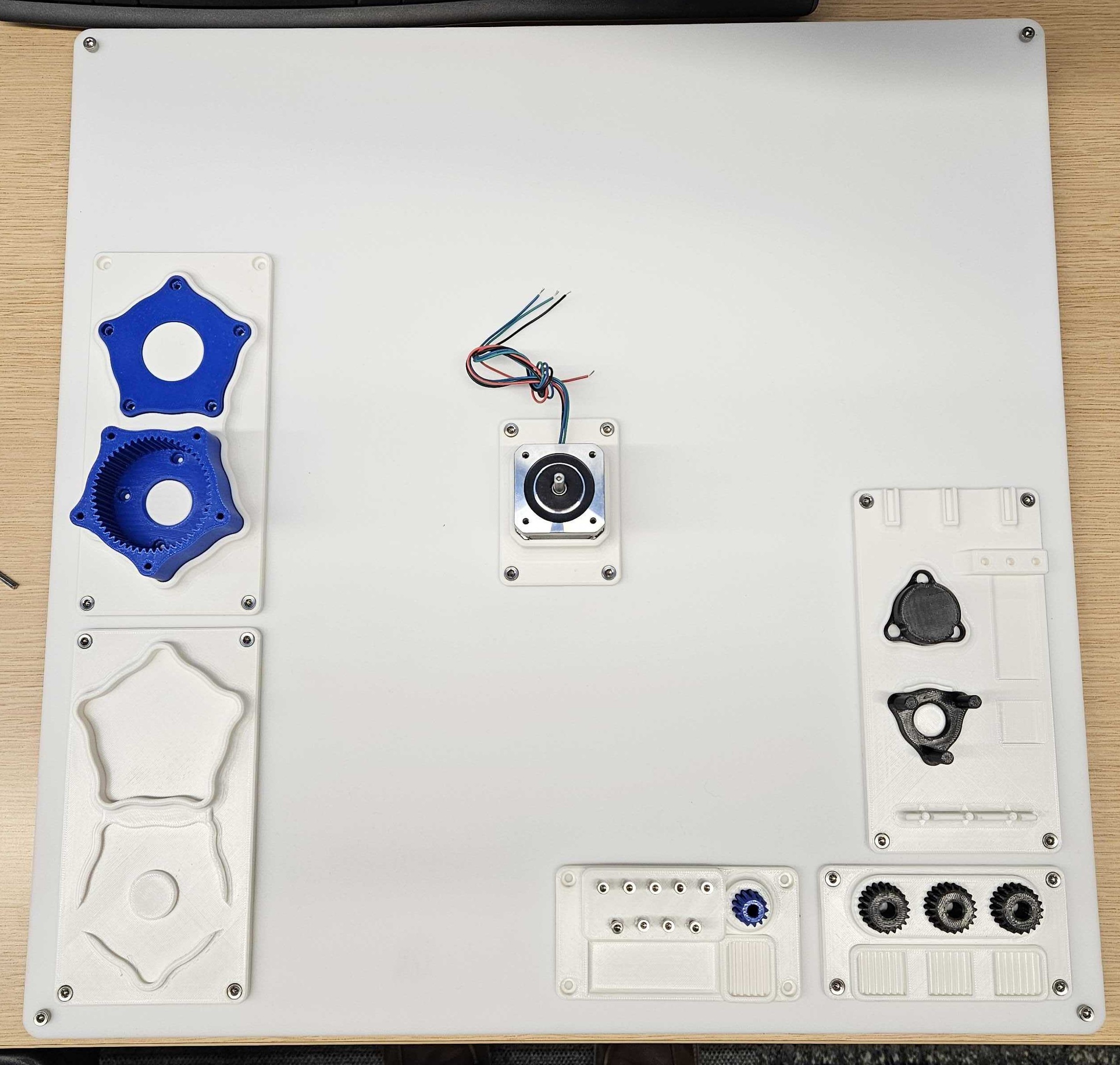}}
\caption{Gearbox assembly benchmarking board (IDB Board~\#3).}
\label{fig:idb_gearbox}
\end{figure}

The gearbox assembly benchmark (IDB Board~\#3) tests a robot's fine manipulation skills in assembling and disassembling a planetary gearbox
driven by a NEMA~17 stepper motor. All of the gearbox
components\footnote{The components of the gearbox include a ring frame, cap, ring, ring base, sun gear, planet gears,
carrier base, carrier cap, and M3 fasteners.} are placed in their respective cutout holders on the board. 
The robot must pick parts from their holders, align gear meshes, and
tighten bolts in the correct sequence. Fig.~\ref{fig:idb_gearbox} shows the board in its
starting state.

The correct sequence to successfully assemble the gearbox and complete the task is as follows: 
\begin{enumerate}
    \item Mount the ring frame onto the stepper motor with four M3$\times$8 bolts.
    \item Insert the sun gear onto the motor shaft (D-spline alignment).
    \item Place the carrier base.
    \item Seat three planet gears, ensuring mesh with both the sun and ring gears.
    \item Snap on the carrier cap.
    \item Place the frame cap.
    \item Secure the frame cap with five M3$\times$12 bolts.
\end{enumerate}
For scoring, points are awarded for sub-tasks related to placement and fastening tasks. 
One point is awarded for each of the following criteria:
\begin{itemize}
    \item All bolts tightened such that they cannot be rotated by hand.
    \item All planet gears meshing correctly.
    \item The carrier assembly rotating freely after completion.
\end{itemize}

\subsection{Datacenter Benchmark: Baseline Task}\label{sec:baseline_task}
Of the three benchmark boards, we select the IDB Board~\#1 cable-cleaning and re-insertion task on the SC~simplex connector as the baseline for this study. This task is representative of the board's core challenge: it combines free-space reaching with contact-rich, sub-millimeter alignment against a fixed receptacle. Concentrating on a single, fully instrumented task lets us hold the protocol constant while ablating six sensor and policy configurations under identical conditions, isolating the effect of each modality rather than confounding it with task variation.

\begin{figure}[htbp]
\centering
\includegraphics[width=0.95\columnwidth]{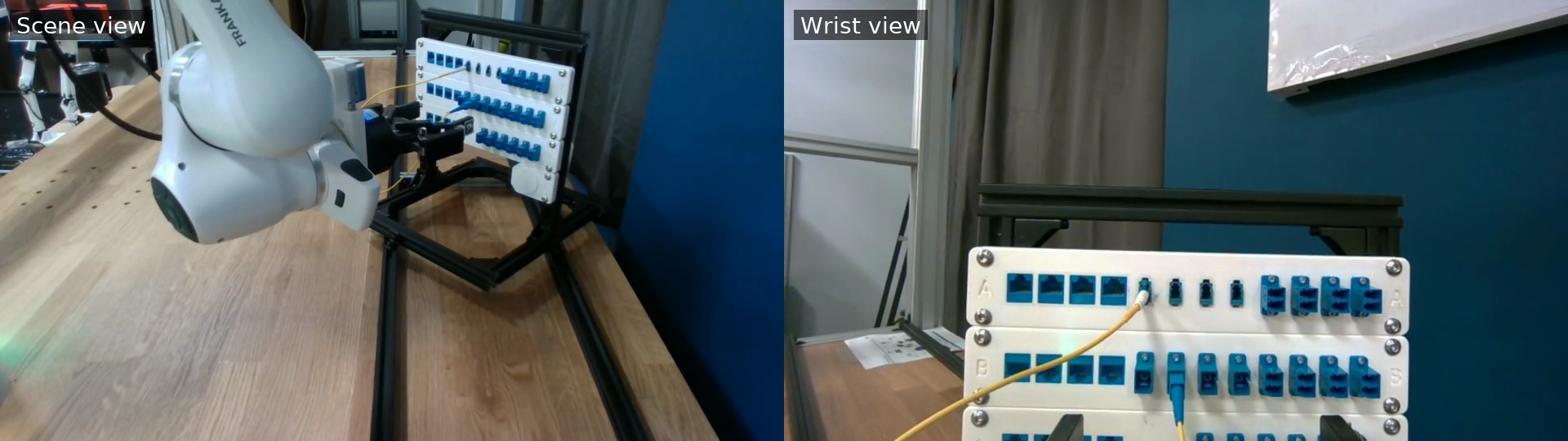}\\[2pt]
{\footnotesize (a) \textit{Grasp}: gripper grasps the connector and
removes it from the receptacle.}\\[4pt]
\includegraphics[width=0.95\columnwidth]{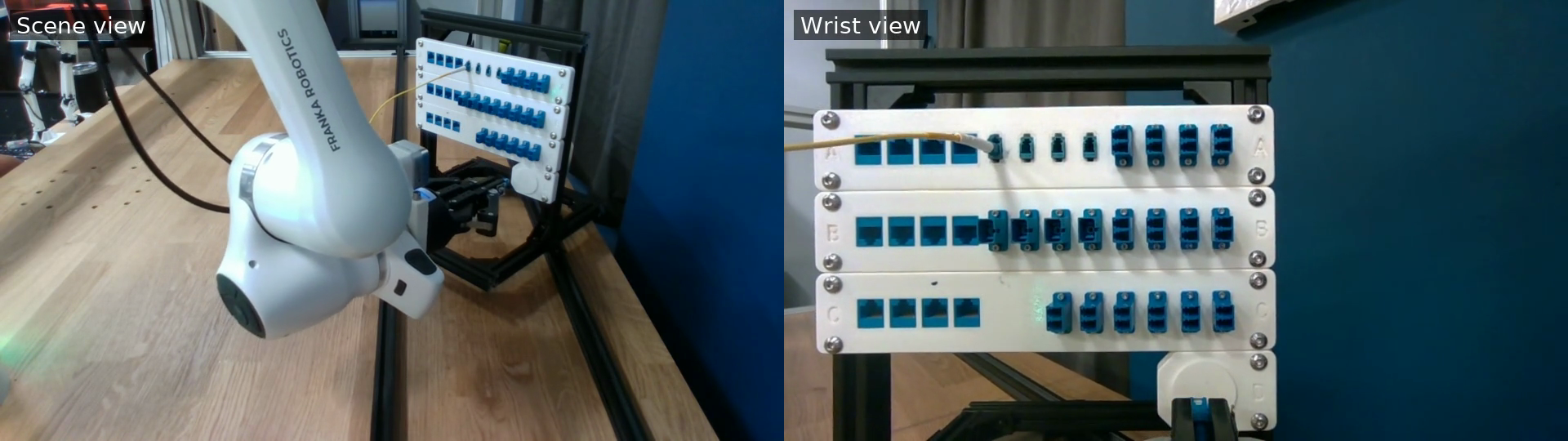}\\[2pt]
{\footnotesize (b) \textit{Clean}: the end of the connector is cleaned against
the cleaning fixture.}\\[4pt]
\includegraphics[width=0.95\columnwidth]{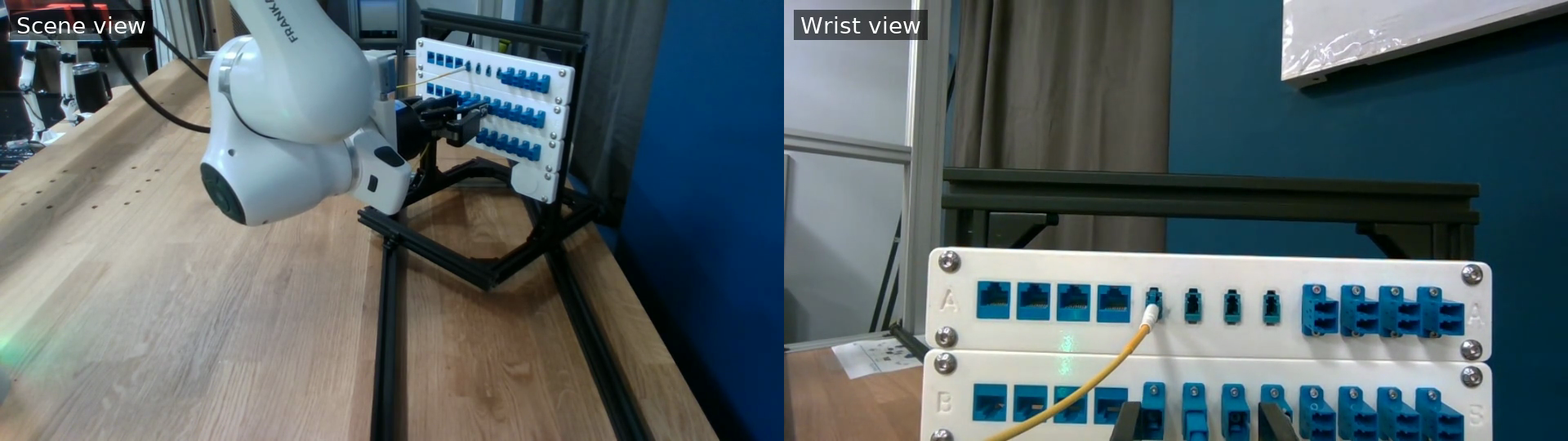}\\[2pt]
{\footnotesize (c) \textit{Insert}: the connector is aligned and reinserted in the receptacle at its original position.}
\caption{The phases of the IDB Board~\#1 baseline task, captured during
a teleoperated demonstration. Each panel shows the scene camera view (left)
and the wrist camera view (right), which are used as policy inputs.}
\label{fig:phases}
\end{figure}

All trials are run at the board's lowest difficulty setting: every port is left empty except the single target connector to be cleaned and re-inserted, so the policy is evaluated on the grasp–clean–insert motion in isolation, without neighboring cables to avoid or occlude the workspace. This choice also reflects the end-effector used in this study: a simple two-finger parallel gripper (Robotiq~2F-85), which lacks the dexterity to manipulate a connector cleanly among tightly packed neighbors, so a sparsely populated board is the appropriate operating regime for it. The task is broken into three phases: \textit{Grasp}\footnote{The robot approaches the plugged connector and closes the gripper to grasp it.}, \textit{Clean}\footnote{The robot moves the connector and presses it against the cleaning pad.}, and \textit{Insert}\footnote{The robot inserts the connector back into its original slot.}. Fig.~\ref{fig:phases} shows the scene and wrist-camera views from a teleoperated demonstration of each phase.

\subsection{Sensor and Hardware Configurations}
Six configurations were evaluated to quantify how different sensor inputs affect task success rate. 
Table~\ref{tab:sensors} summarizes the cameras used in the six configurations. 
The top sub-table lists the three cameras
used: an Intel RealSense D405 mounted on the wrist as the
egocentric baseline in every configuration, an Intel RealSense
D435i providing the industry-standard scene-level RGB and stereo point
cloud, and the EVAL-ADTF3175 direct time-of-flight head from Analog Devices providing
a dense time-of-flight (ToF) point cloud and a 16-bit
infrared (IR) amplitude image (no RGB). The two scene cameras, the EVAL-ADTF3175 and D435i, are compared head-to-head. 
The bottom sub-table enumerates the six configurations, with configuration~6 (RGB wrist only) 
serving as the DP single-camera baseline.

\begin{table}[htbp]
\caption{Cameras and sensor configurations evaluated on the IDB
Board~\#1 datacenter cable-cleaning task.}
\label{tab:sensors}
\begin{center}
\renewcommand{\arraystretch}{1.15}
\setlength{\tabcolsep}{4pt}
\begin{tabular}{p{0.30\columnwidth} p{0.30\columnwidth} p{0.30\columnwidth}}
\toprule
\multicolumn{3}{c}{\textbf{Cameras}} \\
\midrule
\textbf{Sensor} & \textbf{Principle} & \textbf{Modalities} \\
RS D405 (wrist) & Passive IR stereo & RGB + stereo PC \\
RS D435i (scene) & Passive IR stereo & RGB + stereo PC \\
EVAL-ADTF3175 (scene) & Indirect ToF & ToF PC + 16\,b IR \\
\bottomrule
\end{tabular}

\vspace{0.5em}
\begin{tabular}{c p{0.65\columnwidth}}
\toprule
\multicolumn{2}{c}{\textbf{Configurations evaluated}} \\
\midrule
\textbf{\#} & \textbf{Wrist sensor / scene sensor and modalities} \\
1 & RGB Wrist (D405) + RGB Scene (D435i) \\
2 & RGB Wrist (D405) + EVAL-ADTF3175 ToF PC \\
3 & RGB + PC Wrist (D405) \\
4 & RGB Wrist (D405) + RealSense PC Scene \\
5 & PC Wrist only (D405, egocentric) \\
6 & RGB Wrist only (D405) --- DP baseline \\
\bottomrule
\end{tabular}
\end{center}
\end{table}

\section{Imitation Learning Infrastructure}\label{sec:dagros}
\begin{figure}[htbp]
\centerline{\includegraphics[width=0.95\columnwidth]{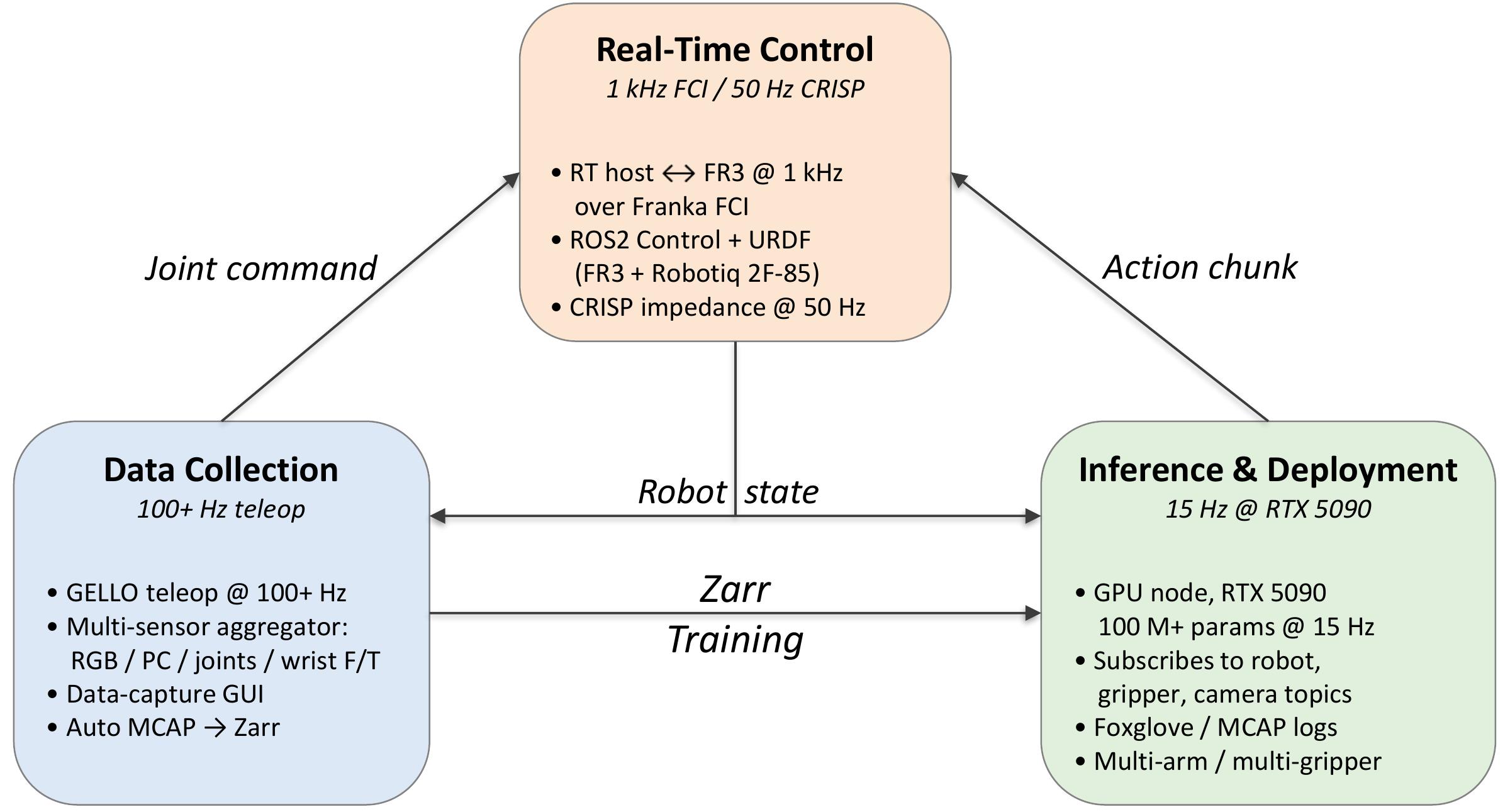}}
\caption{Imitation-learning infrastructure, comprising three subsystems
(data collection, real-time control, and inference \& deployment) connected
by labeled data-flow arrows. Subsystem details and the role of each arrow
are described in Section~\ref{sec:dagros}.}
\label{fig:dagros}
\end{figure}

To support a rapid data collection pipeline for our various hardware and sensor configurations, we developed a scalable imitation learning framework built on
ROS2~\cite{macenski2022ros2} and deployed as modular Docker containers.
Fig.~\ref{fig:dagros} illustrates the system as three subsystems connected
by data flow: a \emph{data collection} subsystem in which a GELLO-style leader arm
\cite{wu2024gello} publishes target joint commands while a data aggregator produces time-aligned RGB, point cloud, joint, and wrist
force/torque (F/T) data streams at 15\,Hz; a \emph{real-time control} subsystem which feeds the target joints 
into the CRISP \cite{pro2025crispcompliantros2} impedance controller running at 50\,Hz, 
whose output drives a Franka FR3 at 1\,kHz via the Franka Control Interface (FCI);
and an \emph{inference and deployment} subsystem on a dedicated GPU node
(RTX 5090) that subscribes to the robot, gripper, and camera topics and
publishes the action chunk outputs of the trained policy at 15\,Hz. 
The arrows connecting the subsystems in Fig.~\ref{fig:dagros} highlight two sets of data flows. 
During training-time data collection, joint commands from teleoperation demonstrations stream 
into the FR3 real-time control subsystem, and the resulting Zarr datasets feed the offline 
model training in the inference and deployment subsystem. 
During deployment, the action chunks output by the trained policy feed the real-time control subsystem, 
and robot state is published back to the inference and deployment subsystem.

\subsection{Implementation Details}
A few key aspects of the implementation are worth highlighting here. 
The data collection subsystem uses an intuitively configurable data aggregator with a master data stream that all
other data streams align to during collection. Also included is a data-capture GUI, allowing the operator to start, stop, 
and pause recording without leaving the teleoperation loop. Once the MCAP dataset is captured, a validity checker is available
to check the dataset for correctness and explicitly raise issues to the user before
an automated workflow runs to transform the dataset to the training-ready Zarr
format. The real-time control subsystem uses an Ubuntu 24.04 RT-kernel host with ROS2 Control and URDF
integration to drive the final robot assembly composed of the FR3, Robotiq 2F-85 gripper, and wrist-mounted RealSense D405. 
All demonstrations and deployment runs are logged in MCAP format and visualized with Foxglove for offline analysis. 
% DAG-ROS is designed for straightforward integration of robot arms from different vendors and bimanual configurations, 
% requiring only that the vendor supply appropriate ROS control packages and URDFs. 
% This way any imitation learning project can reuse the DAG-ROS infrastructure layer 
% to reduce the robot bring-up overhead.

\section{AG-iDP3: Multimodal Diffusion Policy}\label{sec:agidp3}
\begin{figure}[htbp]
\centerline{\includegraphics[width=0.95\columnwidth]{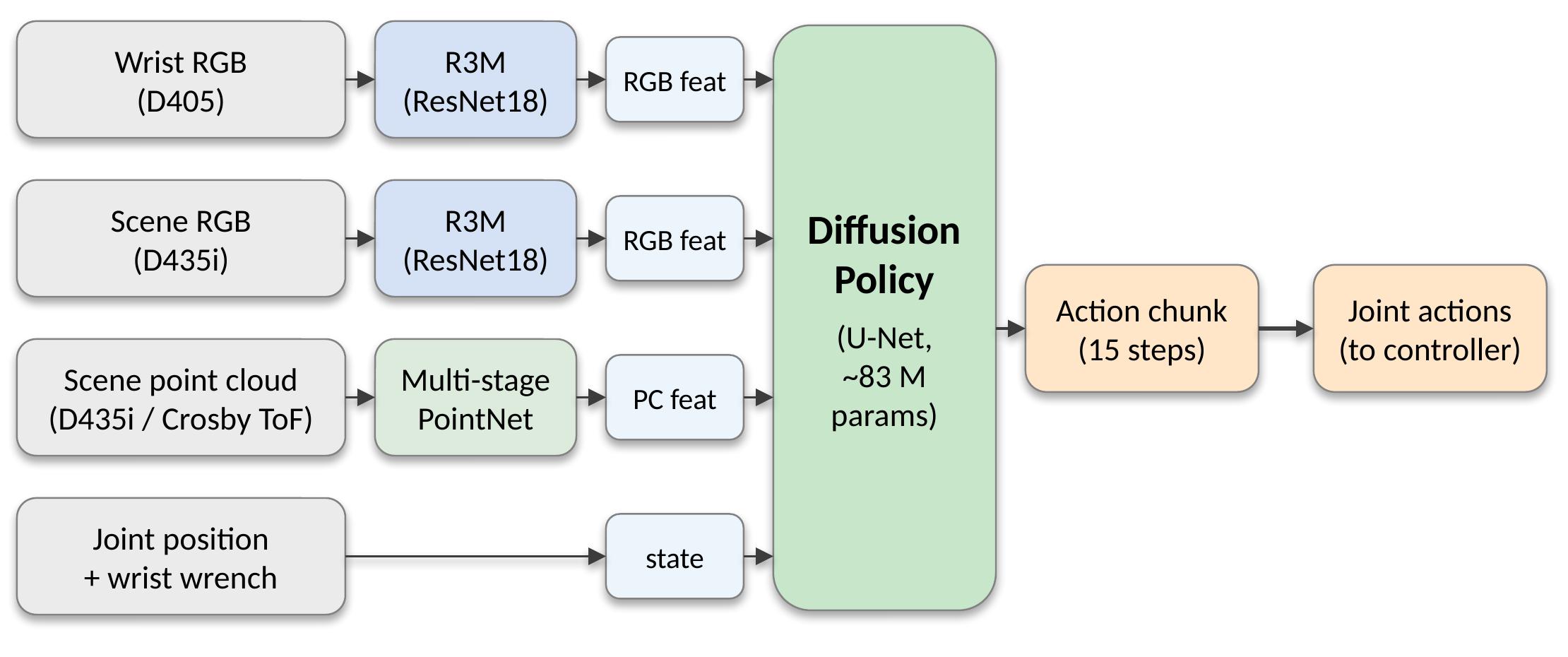}}
\caption{AG-iDP3 framework, shown in its full multimodal configuration. 
Wrist RGB (D405) and the optional scene RGB (D435i) are each encoded by an R3M
\cite{nair2022r3m} ResNet18 backbone; the scene point cloud
(D435i stereo or EVAL-ADTF3175 ToF) is encoded by a multi-stage PointNet
\cite{qi2017pointnet}; proprioceptive states (joint position and
wrist wrench) are unencoded. All vectors are concatenated into a single 
observation that is input to a diffusion U-Net which outputs an action chunk of $T$ joint actions.}
\label{fig:agidp3}
\end{figure}

At the core of the system is \textit{AG-iDP3}, a multimodal diffusion-based policy framework derived from existing work done on 3-D Diffusion Policy (DP3
\cite{ze2024dp3}, iDP3 \cite{ze2025idp3}) and the broader diffusion-policy
formulation \cite{chi2023diffusion,ho2020ddpm}. Fig.~\ref{fig:agidp3}
shows the data flow through the policy. RGB images from the wrist-mounted RealSense D405 and,
optionally, the scene RealSense D435i are each encoded by an
R3M \cite{nair2022r3m} backbone (ResNet18
\cite{he2016resnet}) into a per-image RGB feature
vector. AG-iDP3 supports both a frozen and a fine-tuned R3M backbone; in this work, fine-tuning on the collected task demonstrations yielded better performance and is used in all reported results. 

The scene point cloud data from both the D435i and the EVAL-ADTF3175 ToF cameras is downsampled and encoded by a lightweight multi-stage PointNet
\cite{qi2017pointnet} encoder. The proprioceptive states, composed of
joint positions and wrist wrench, are unencoded and pass through directly as the state vectors. All vectors are concatenated into a single observation vector. During the inference step, the combined observation vector is passed to a diffusion U-Net which produces an action chunk with the next $T$ predicted
joint actions. Chunks from successive inference steps are blended via temporal ensembling and converted to a smooth 50\,Hz command stream (Section~\ref{sec:motion}).  The chunk length $T$ is configurable in AG-iDP3; in this work we use $T=15$.

\subsection{Model Characteristics}
We evaluate four model architectures within the AG-iDP3 framework; Table~\ref{tab:params}
reports the parameter footprint of each, broken down by encoder and U-Net contribution.
The point-cloud-only variant has the smallest footprint (68.77\,M) thanks to the lightweight multi-stage PointNet
\cite{qi2017pointnet} encoder, but lacks RGB texture. Adding a single-camera RGB input
with an R3M encoder grows the model by 20\,M parameters (89.25\,M total).
Adding a second RGB input doubles the R3M cost and grows
the U-Net to 92.58\,M parameters (114.93\,M total). The combined point-cloud
+ single-camera RGB architecture sits between these extremes,
adding only $\sim$5\,M parameters over the single-camera RGB variant
while preserving 3-D awareness through the PointNet branch.

\textbf{Architecture choice.} AG-iDP3 retains the convolutional diffusion U-Net of iDP3~\cite{ze2025idp3} for two reasons. First, it keeps the backbone fixed across all six configurations so that our results isolate the effect of input modality rather than confounding it with a change of policy architecture. Second, the convolutional formulation is the better-suited choice in our data regime, as prior work reports that the U-Net variant matches its transformer counterpart on most tasks while being substantially less sensitive to hyperparameters, and recommends it as the default when demonstration episodes are limited~\cite{chi2023diffusion}. With roughly 100 demonstrations per phase, this data efficiency is the binding constraint for on-site industrial data collection, where transformer policies typically require larger demonstration sets to reach comparable performance. 

\begin{table}[htbp]
\caption{Parameter footprint by model variant.}
\label{tab:params}
\begin{center}
\renewcommand{\arraystretch}{1.15}
\begin{tabular}{lcccc}
\toprule
Config & PC Enc.\ & R3M Enc.\ & U-Net & Total \\
\midrule
iDP3 (PC only)            & 0.27\,M & ---       & 68.5\,M & 68.8\,M \\
DP (1$\times$ RGB)        & ---     & 11.2\,M  & 78.1\,M & 89.3\,M \\
DP (2$\times$ RGB)        & ---     & 22.4\,M  & 92.6\,M & 114.9\,M \\
AG-iDP3 (PC+RGB) & 0.27\,M & 11.2\,M  & 83.2\,M & \textbf{94.6\,M} \\
\bottomrule
\end{tabular}
\end{center}
\end{table}

\subsection{Per-Phase Wrench Gating}
The modality-gating mechanism, which selects which sensor inputs feed
each diffusion-policy model, is exercised at the per-phase level for
the wrist wrench input. Across the three phases of the behavior tree
(Section~\ref{sec:bt}), we found that wrench input only helped during
the contact-rich \emph{insert} phase; for the \emph{grasp} and
\emph{clean} phases, including wrench provided no measurable benefit
and was therefore gated off in those phases' trained policies. This
per-phase configuration is enabled directly by the framework's
input-gating architecture and illustrates how the same model template
can be specialized to the contact characteristics of each phase
without architectural changes.

\section{Motion Control Deployment} \label{sec:motion}
\begin{figure}[htbp]
\centerline{\includegraphics[width=0.95\columnwidth]{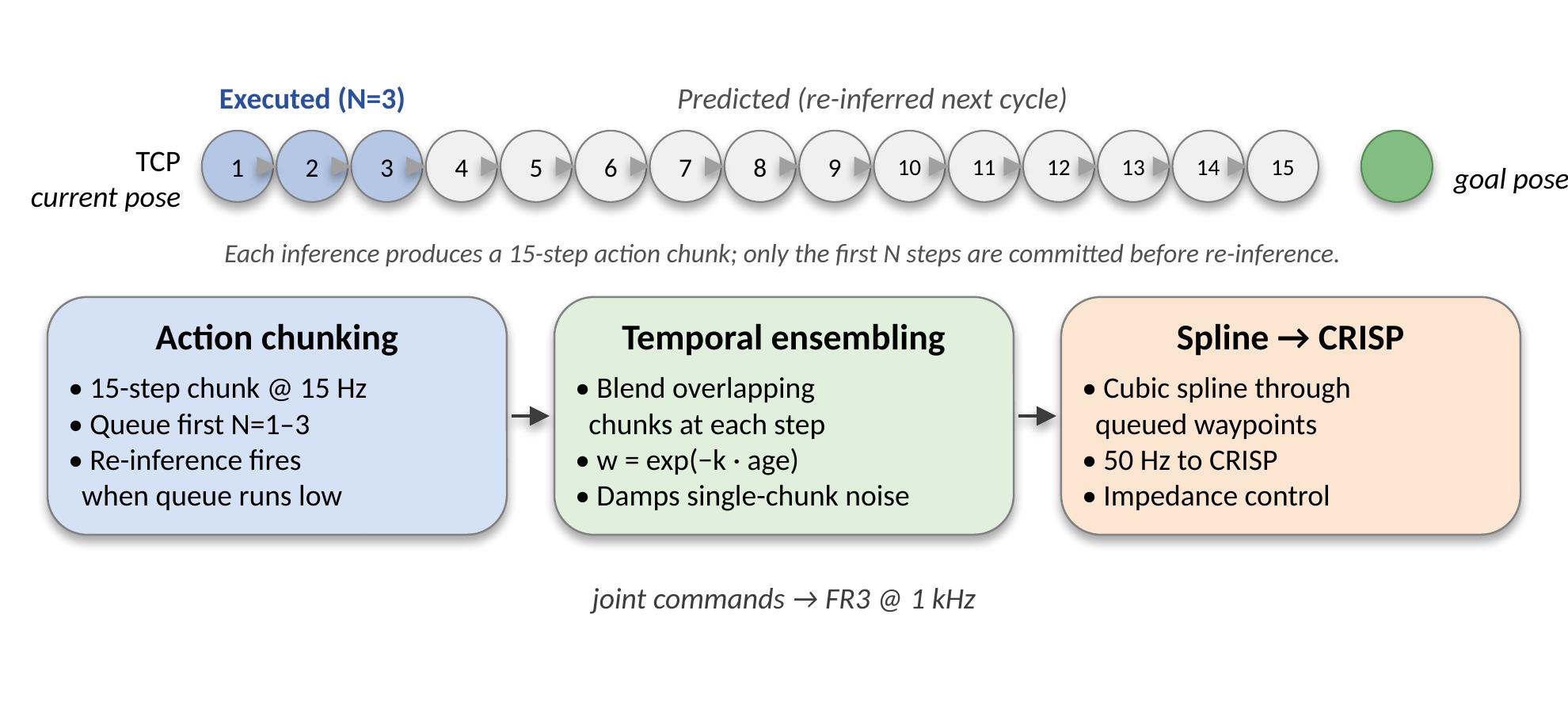}}
\caption{Each inference produces a
15-step action chunk (top), from which only the first $N$ steps are
committed to execution before the next inference step begins. Overlapping
predictions across chunks are temporally ensembled
(Eq.~\ref{eq:tempens}) and an interpolated cubic spline derives commands through the committed action steps for 50\,Hz streaming to the CRISP impedance controller.}
\label{fig:motion}
\end{figure}

AG-iDP3 outputs are integrated into a real-time control framework that
converts the 15-step action chunks produced by inference into smooth, compliant robot joint commands.
Fig.~\ref{fig:motion} shows the deployment pipeline. The top part of the figure
visualizes a single inference's 15-step action chunk where the first three highlighted 
steps are sent to the controller, while the remaining twelve are blended with predictions from the next inference via temporal ensembling. 
The bottom part of the figure decomposes the post-inference processing
into three stages: action chunking, temporal ensembling, and a cubic
spline interpolation that streams commands at 50\,Hz to the CRISP impedance
controller. Together, these stages turn discrete predictions into
smooth, contact-aware motion.

\ifralshort

The first $N=3$ steps of each action chunk are committed for execution while a background re-inference fires adaptively when the queue runs low, decoupling inference latency from controller throughput. A cubic spline anchored at the last executed action interpolates through the committed waypoints and is sampled at 50\,Hz to produce a dense joint-space command stream for the CRISP impedance controller. This interpolation upsamples the 15\,Hz policy rate to the 50\,Hz control rate without exceeding velocity limits, and eliminates the known boundary discontinuity at chunk transitions where the first predicted action is further from the current TCP than inter-action spacing within a chunk due to the discrepancy between commanded and executed trajectories \cite{hhe_dual_tracking_RAL2024}. Successive inferences produce overlapping predictions for the same future timesteps~\cite{zhao2023act}; we blend them via temporal ensembling \label{sec:temporal_ensembling} with exponentially decaying weights:
\begin{equation}
\bar{a}^{(t)} = \frac{\sum_i w_i\, a_i^{(t)}}{\sum_i w_i},\qquad
w_i = \exp(-k \cdot \mathrm{age}_i),
\label{eq:tempens}
\end{equation}
where $k=0.01$ damps inter-chunk noise while closely tracking the most recent inference.

\else
\subsection{Trajectory Execution (Action Chunking)}
The first $N$ steps of each action chunk are queued for execution while
a background re-inference fires when the queue runs low, decoupling
inference latency from controller throughput. The re-inference trigger 
time is set adaptively by tracking the execution latency of recent inference steps, 
so that a new chunk arrives just before the queue empties. The number of committed
steps $N$ is configurable; in this work we use $N=3$, which balances
reactive behavior (smaller $N$ requires more frequent inference updates) with
smooth motion (larger $N$ reduces the frequency of redirection). The
remaining queued steps are blended with the next inference's
predictions via temporal ensembling (Section~\ref{sec:temporal_ensembling}).

\subsection{Trajectory Smoothing}
The interpolation step fits a smooth trajectory starting at the last
executed action and passing through the next $N$ predicted waypoints,
then samples it at the controller frequency (50\,Hz) to produce a
dense stream of joint-space targets for the CRISP impedance
controller. In joint-space mode, each arm joint is fit with its own
cubic spline using arc-length time parameterization, giving each
joint constant velocity along its individual
path, while the gripper is interpolated piecewise-linearly. In
Cartesian-space mode (used by some configurations of AG-iDP3 but
not the cable-cleaning task reported here), positions are
interpolated by cubic splines, orientations by SLERP using a shared
arc-length parameterization, and the resulting Cartesian targets are
converted to joint-space via iterative inverse kinematics before
being streamed to the controller.

This interpolation step serves two purposes. First, the policy
predicts actions at the 15\,Hz inference rate at which demonstrations
were collected, but the impedance controller expects 50\,Hz
setpoints; sending raw inference outputs would force the controller
to traverse each inter-waypoint segment in 1/15\,s, exceeding
velocity and acceleration limits. Sampling the spline at 50\,Hz yields a $\sim$3$\times$ denser
command stream that respects these limits. 
Second, action-chunking policies exhibit a known boundary discontinuity at the start of each chunk: 
the first predicted action is typically further from the current TCP than the spacing between subsequent actions within the chunk due to the discrepancy between commanded and executed trajectory \cite{hhe_tracking_icra2023,hhe_dual_tracking_RAL2024}. 
Anchoring the spline at the last executed action and resampling at 50\,Hz spreads this larger first-action gap across multiple interpolated steps, eliminating per-chunk oscillation in the robot motion.

\subsection{Temporal Ensembling} \label{sec:temporal_ensembling}
Successive inferences produce overlapping predictions for the same
future timesteps~\cite{zhao2023act}.  We use temporal ensembling to
blend these overlapping predictions with more recent inferences
weighted more heavily. Let $a_i^{(t)}$ denote the action predicted at
inference~$i$ for execution time~$t$, and let $\mathrm{age}_i$ be the
elapsed time since inference~$i$. The blended action is the
exponentially weighted average
\begin{equation}
\bar{a}^{(t)} = \frac{\sum_i w_i\, a_i^{(t)}}{\sum_i w_i},\qquad
w_i = \exp(-k \cdot \mathrm{age}_i),
\label{eq:tempens}
\end{equation}
where $k>0$ controls how quickly older predictions decay. Larger
$k$ favors the most recent inference and yields more reactive
behavior; smaller $k$ produces smoother motion at the cost of some
lag relative to the latest inference. In AG-iDP3, $k$ is configurable;
in this work we use $k=0.01$, which damps the single-chunk noise that
would otherwise appear at chunk boundaries while still closely
tracking the most recent inference.

\fi

\section{Behavior Tree Architecture}\label{sec:bt}
\begin{figure}[htbp]
\centerline{\includegraphics[width=0.95\columnwidth]{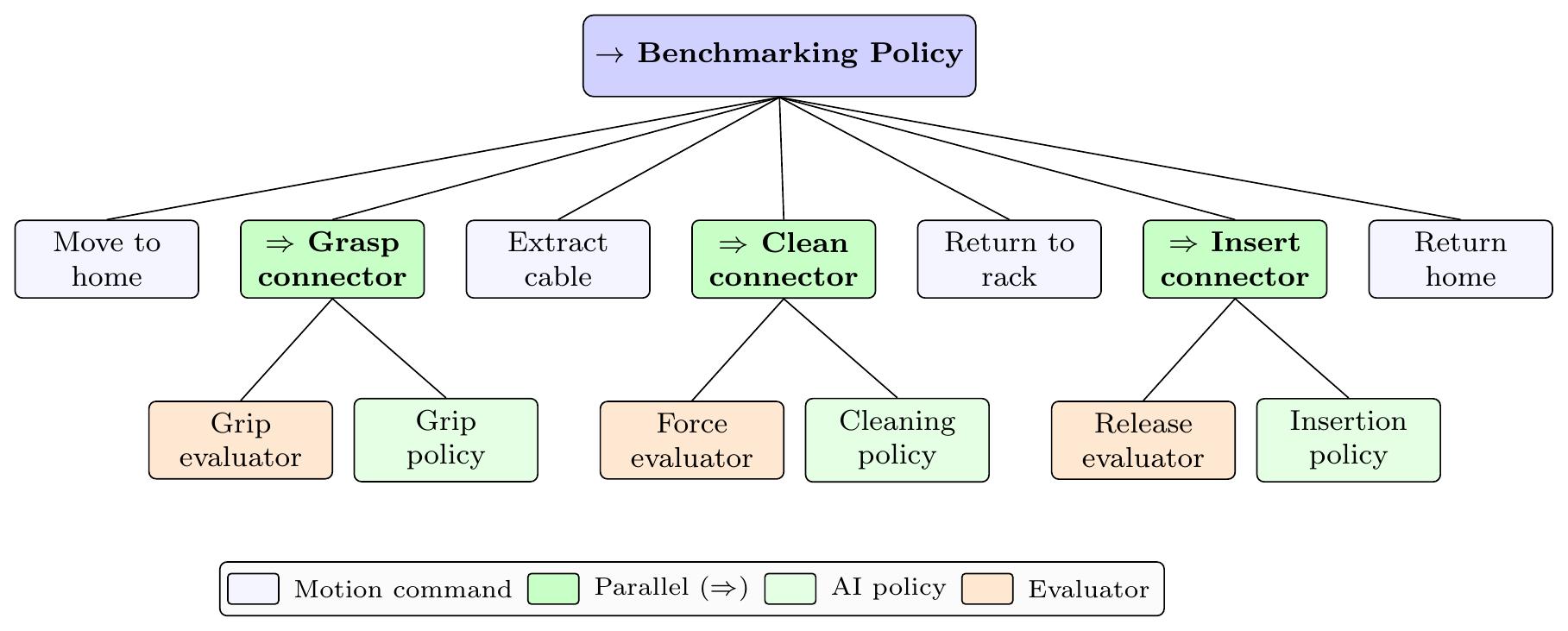}}
\caption{Hybrid behavior tree orchestrating the datacenter
cable-cleaning task.}
\label{fig:bt}
\end{figure}

Fig.~\ref{fig:bt} shows the PyTrees behavior
tree~\cite{colledanchise2018bt} orchestrator that defines the cable-cleaning benchmarking policy flow. The root \textsc{Sequence} ($\rightarrow$)
node executes its children
left-to-right. The subnodes of the root node (\emph{move to home},
\emph{grasp connector}, \emph{extract cable}, \emph{clean connector},
\emph{return to rack}, \emph{insert connector}, and
\emph{return home}) comprise the full cable-cleaning policy. 

Three of the root node's children are \textsc{Parallel} ($\Rightarrow$)
nodes, each containing a policy node and an evaluator node that run
concurrently. The evaluator is necessary because AI policies do not
terminate on their own, as successive inferences keep generating new
action chunks indefinitely. When the evaluator returns success, a stop
request is sent to the policy node before the parent \textsc{Parallel}
node itself returns success.

\begin{enumerate}
    \item \textbf{Grasp}---The AI policy locates and grips the connector on the rack. The evaluator detects when the gripper position drops below a closure threshold for a preset time, or after 1 minute timeout.
    \item \textbf{Clean}---The AI policy wipes the connector face against the cleaning pad. The evaluator detects when the contact wrench magnitude exceeds a threshold for a preset time, or after 1 minute timeout.
    \item \textbf{Insert}---The AI policy aligns the connector with the receptacle, inserts it, then opens the gripper. The evaluator detects when the gripper position rises above an opening threshold for a preset time, or after 1 minute timeout.
\end{enumerate}

This hybrid architecture combines learned policies for dexterous manipulation with
classical motion primitives for reliable and safe movements,
yielding a system that balances adaptability and reliability by applying each method where it is most appropriate for the phase.

\section{Results}\label{sec:results}
We evaluated the six policy configurations shown in Table~\ref{tab:sensors}
head-to-head with the orchestrator shown in Fig. 9 on the IDB Board~\#1 datacenter cable-cleaning and re-insertion benchmark task. All four SC simplex receptacle positions were tested 12 times each, for a total of 48 trials per
configuration. A success is conditioned on the previous task's success, for example, insertion success rate is defined as $(\text{Insert Success}) / (\text{Clean Success})$. The total score is defined as
$(\text{Grasp Success}+\text{Insert Success})/96$. The cleaning phase is excluded from the total
score because it succeeded on 100\% of trials across every configuration after a successful grasping.

\subsection{Key Findings}
\begin{figure}[htbp]
\centerline{\includegraphics[width=0.95\columnwidth]{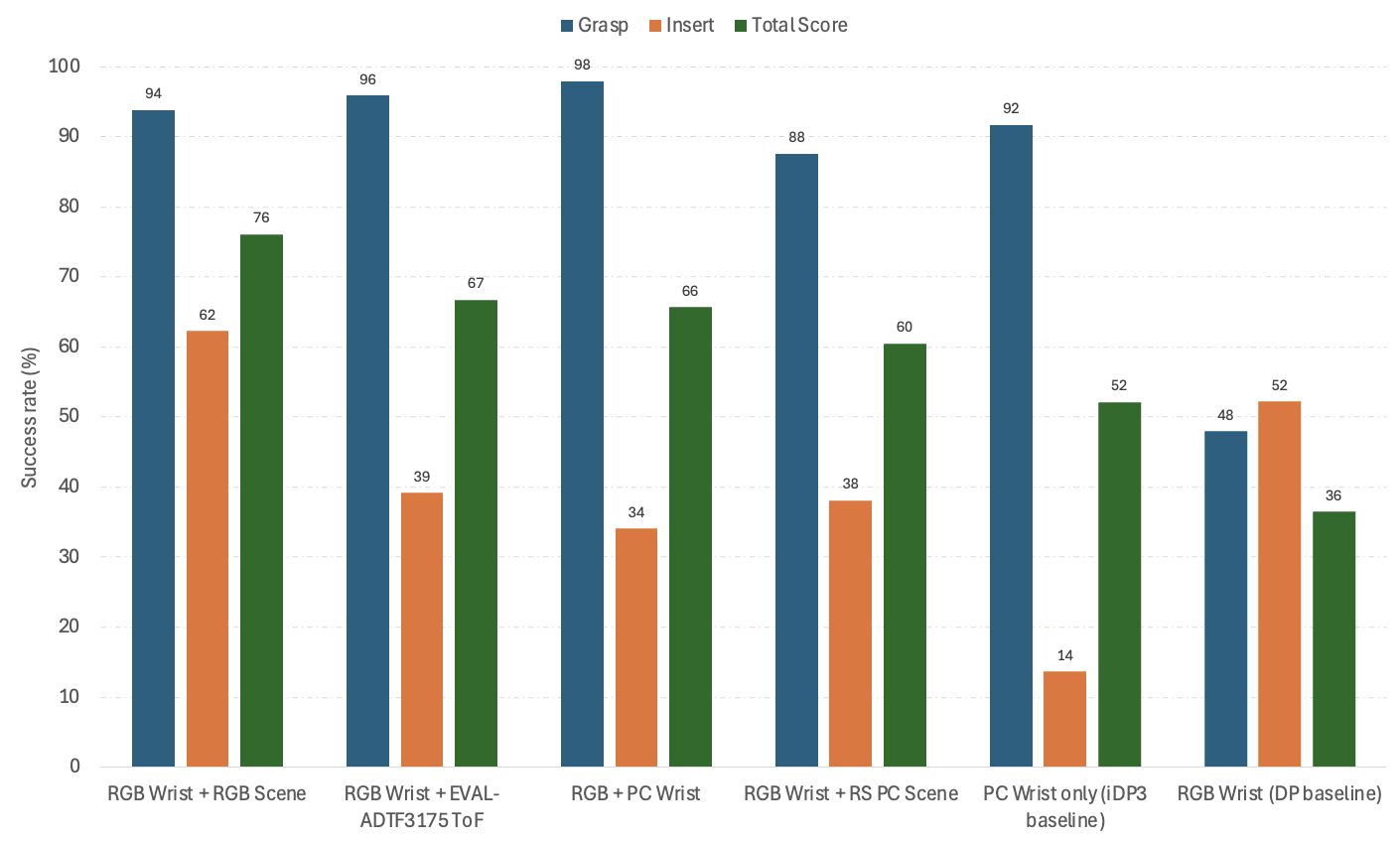}}
\caption{Grasp, insert, and total success rates across the six
configurations of Table~\ref{tab:sensors} (48 trials per
configuration). The best multimodal expansion (RGB Wrist + RGB
Scene) with R3M reaches 76\% total versus 36\% for the single-camera
RGB DP baseline. The head-to-head comparison between the two scene cameras shows the EVAL-ADTF3175 ToF outperforming the RealSense D435i by 7~percentage points. Cleaning succeeded on 100\% of trials in every
configuration and is excluded from the total score.}
\label{fig:results}
\end{figure}

Fig.~\ref{fig:results} reports grasp, insert, and total success rates
for each configuration. The key takeaways are:
\begin{itemize}
    \item Whenever the policy has 3-D context available, either through point cloud or through two RGB cameras providing different perspectives, the grasp phase performs well, with grasp success rates ranging from 88\% to 98\% across the five configurations that include 3-D context. Only in the single-RGB configuration (config 6), where 3-D context is entirely removed, does the grasp success rate drop to 48\%.
    \item With the exception of the RGB-only baseline, insertion is the phase that differentiates total scores. Across configurations 1--5, insertion score varies by 48 percentage points while grasp success only varies by 10 percentage points. 
    \item The best total score (76\%) is achieved by the multi-view RGB configuration (RGB wrist + RGB scene) with R3M, which combines the
    strong grasping behavior of wrist RGB with the high-resolution
    scene context needed for fine alignment. 
    \item In the direct head-to-head comparison between the two scene cameras, the EVAL-ADTF3175 ToF outperforms the RealSense D435i by 7~percentage
    points, suggesting that indirect ToF is better suited to industrial
    datacenter conditions.
    \item All four multimodal expansions (configurations 1--4) outperform the single RGB camera DP baseline (config 6, 36\% total), and the egocentric point-cloud-only iDP3 (configuration 5, 52\% total) replicates the iDP3 finding that an egocentric point-cloud policy beats an egocentric RGB policy.
    \item All three phases of the cable-cleaning benchmarking task (grasp, clean, and insert) required roughly 100 demonstrations each to 
    train a well-performing policy across every ablation. At this demonstration count, AG-iDP3-trained policies are practical
    for on-site data collection and re-training whenever the task or
    hardware configuration changes.
\end{itemize}

\ifralshort

\else
\subsection{Additional Considerations}
A few caveats and design-choice notes are worth noting alongside
the key findings.

\textit{Point-cloud resolution vs.\ industrial robustness.} The
resolution of both PC scene cameras, further downsampled in the
PointNet encoder, was not fine enough to resolve the receptacle
features needed for tight-clearance insertion. This is likely why
the dual-RGB configuration outperformed both RGB-wrist + PC-scene
configurations. In an industrial setting, however, where lighting is
not constant, equipment discolors with age and contamination, and
airborne particulates are common, ToF sensing may provide more
reliable output than passive stereo even when the experimental
configuration favored the dual RGB configuration.

\textit{Scoring rationale.} A scoring rule was fixed before
experimentation. Summing grasp and insert successes (and excluding
cleaning, which always succeeded) reflects the operational reality
that a successful insertion is meaningless without a successful
grasp. If the robot fails to grasp and extract the cable, there is
no opportunity for the insertion phase to succeed; if the robot
grasps and cleans but fails to insert, the task is partially
complete.

\textit{Caveat on per-phase comparison for the RGB-only baseline.}
Per-phase comparisons against the RGB-wrist-only configuration should
be interpreted carefully. We observed that ports 2 and 3 had higher
success rates than ports 1 and 4 on both phases. Because the RGB-only
configuration's low grasp success rate filtered out a large fraction
of insert attempts (and grasp is the simpler task, so a failed grasp
likely implies a failed insert), the insert success rate for that
configuration is not directly comparable to those of configurations
that successfully grasp on most trials. For this reason, we should not draw
conclusions about relative configuration performance on the insertion phase 
numbers alone.

\subsection{System-Level Outcome}
The hybrid behavior tree (Fig.~\ref{fig:bt}) chained grasp, clean,
and insert end-to-end on every evaluated configuration, showing that
the parallel pairing of learned policy with classical evaluator
scales beyond a single phase. Compared with the classical pipeline of
Section~\ref{sec:classical} used on the NIST board challenge, deploying the learned policy on a new
task requires only $\sim$100 teleoperated demonstrations, whereas the
classical pipeline demands thousands of labeled images for perception
training, extensive per-stage parameter tuning, and reprogramming of motion planning and force/torque steps
for each new task.

\fi

\subsection{Robustness Observations}
The learned policy was brittle to small visual scene changes. Two
representative failure modes were observed: (i) removing a background
object (a red polymer block) that was present at training time caused
the policy to fail; (ii) changing the routing of a cable along the robot
arm (loose vs.\ tied) also caused failure. 
We mitigated (i) by training and deploying on cropped images so that only the controlled task area is fed to the policy, and (ii) by adopting pulley-based cable management so that the 
cables return to the same position regardless of the path that the arm takes. While these task-specific workarounds restored reliable behavior, 
the underlying robustness issue persists: the visual encoder appears to latch onto incidental scene features rather than
task-relevant ones, motivating data augmentation, domain randomization,
and broader scene diversity in training as direction for future work.

\ifralshort
\section{Conclusion and Future Work}
This work presents IDB Board~\#1, a validated datacenter cable-manipulation benchmark, two preliminary companion board designs, the imitation-learning infrastructure, and the AG-iDP3 policy framework. On the IDB Board~\#1 datacenter cable-cleaning task, a multimodal diffusion policy trained with $\sim$100 demonstrations per phase more than doubles the single-camera baseline (76\% vs.\ 36\% total score), and a hybrid behavior tree combining learned policies with classical motion primitives provides a practical deployment architecture. Future work targets richer encoders (DINOv3, DiT), additional sensing modalities (wrist ToF, tactile, event cameras), bimanual coordination, and edge deployment on NPUs and analog accelerators.

\else
\section{Conclusion}
This work presents a three-part system for industrial dexterous
manipulation: the IDB benchmarking program (three board designs
spanning datacenter cable management, automotive cable harnesses, and
gearbox assembly), the DAG-ROS imitation-learning infrastructure, and
the AG-iDP3 multimodal diffusion policy framework. We compare a modular classical pipeline with the end-to-end learned
alternative and quantify the benefits of the latter, from which three observations stand out. First,
the classical pipeline is functional but fragile: pose-estimation errors,
lighting sensitivity, and the high tuning and reprogramming overhead at every stage
limit how far it scales beyond a controlled bench setup. Second, a
multimodal diffusion policy with $\sim$100 demonstrations per phase
more than doubles the single-camera DP baseline on total task score 
(76\% vs.\ 36\%), and every multimodal policy also outperforms the iDP3 baseline (52\%) that uses only a single egocentric point-cloud view. Third, the
hybrid behavior tree, which pairs learned policies with concurrent evaluators and chains them together with classical motion primitives, provides a practical and flexible deployment that lets the user apply each control method where it is best suited. Together, the IDB Board~\#1 benchmark,
the DAG-ROS infrastructure, and the AG-iDP3 framework form a
self-contained recipe for evaluating dexterous manipulation on
high up-time industrial tasks, and the design is
reproducible from the descriptions and figures in this paper.

\subsection{Future Directions}
Ongoing research focuses on three threads.

\textit{Algorithmic improvements} include richer pre-trained encoders
for point-cloud, depth, and grayscale inputs (e.g., DINOv3
\cite{simeoni2025dinov3}, DiT \cite{peebles2023dit}),
multi-task or semantic encoding, dynamic input resolution, dynamic
prediction horizons, Cartesian-space action models, and a slow-fast
policy that pairs a high-rate reactive controller with a low-rate
planner. Robustness and data scaling can be addressed using domain
randomization, simulation-based training with data augmentation, and
reinforcement-learning fine-tuning.

\textit{Sensing and platform extensions} include direct head-to-head
evaluation of an ADI Tembin wrist ToF sensor against the D405; event
cameras for low-latency motion cues; tactile sensing for local
contact and slip detection; an ADI 6-DOF wrench sensor compared with
the FR3's joint-torque-derived wrench; multi-arm bimanual coordination
for routing, fixturing, and handoff; dexterous multi-finger hands for
in-hand reorientation; and haptic teleoperation devices for richer
contact-rich demonstrations.

\textit{Compute and deployment} improvements target faster inference, higher
action rates, lower latency, and better power efficiency. These goals can be addressed by moving from RTX-class workstations to
edge deployment on neural processing units (NPUs) and analog
neural-network accelerators.
\fi

\ifdoubleblind
\else
\section*{Acknowledgment}\label{sec:ack}
The authors thank the many collaborators who contributed to this work.
The DAG-ROS infrastructure was led by Max Lehuraux, with contributions from DexAI Group including
Tao Yu and Jame Sun. Project management was led by Ashish Rainu. Hardware setup and integration were handled by Audren Cloitre,
and the prototyping team led by Hugo Emidio supported fabrication across the IDB program. The IDB benchmarking boards were designed by
Connor Redding and Dan Barkus. Secondary-site testing
was carried out by Matt Vancleave and Steven Dang. Simulation consulting and support were provided by
Philip Sharos,  Wonju Lee, and Matteo Grimaldi. The authors also thank
Yuval Zukerman for management and consulting support
throughout the project.

All three benchmarking board designs are open source at 
\url{https://github.com/analogdevicesinc/ag-industrial-dexterity-benchmarks}
as of the time of this writing. 
% The codebase of \textit{DAG-ROS} and \textit{AG-iDP3} is currently private; interested readers are
% encouraged to contact the authors with questions or to request access.
\fi

\bibliographystyle{ieeetr}
\bibliography{mybib}

\end{document}